\documentclass[conference]{IEEEtran}
\IEEEoverridecommandlockouts
\usepackage{cite}
\usepackage{amsmath,amssymb,amsfonts}
\usepackage{algorithmic}
\usepackage{graphicx}
\usepackage{textcomp}
\usepackage{xcolor}
\usepackage{hyperref}
\def\BibTeX{{\rm B\kern-.05em{\sc i\kern-.025em b}\kern-.08em
    T\kern-.1667em\lower.7ex\hbox{E}\kern-.125emX}}
\begin{document}

\title{Knowledge Graph Guided Semantic Evaluation of Language Models For User Trust}
\makeatletter
\newcommand{\linebreakand}{%
  \end{@IEEEauthorhalign}
  \hfill\mbox{}\par
  \mbox{}\hfill\begin{@IEEEauthorhalign}
}
\makeatother

\author{\and
\IEEEauthorblockN{\textbf{Kaushik Roy}}
\IEEEauthorblockA{Artificial Intelligence Institute\\
University of South Carolina\\
Columbia, SC, USA \\
\texttt{kaushikr@email.sc.edu}}
\and
\IEEEauthorblockN{\textbf{Tarun Garg}}
\IEEEauthorblockA{Birla Institute of Technology\\
Pilani \\
\texttt{f20160450h@alumni.bits-pilani.ac.in}}
\linebreakand
\IEEEauthorblockN{\textbf{Vedant Palit}}
\IEEEauthorblockA{Indian Institute of Technology \\
Kharagpur \\
\texttt{vedantpalit@kgpian.iitkgp.ac.in}} \\
\and
\IEEEauthorblockN{\textbf{Yuxin Zi}}
\IEEEauthorblockA{Artificial Intelligence Institute\\
University of South Carolina\\
Columbia, SC, USA \\
\texttt{yzi@email.sc.edu}}
\linebreakand
\IEEEauthorblockN{\textbf{Vignesh Narayanan}}
\IEEEauthorblockA{Artificial Intelligence Institute\\
University of South Carolina\\
Columbia, SC, USA \\
\texttt{vignar@sc.edu}} \\
\and
\IEEEauthorblockN{\textbf{Amit Sheth}}
\IEEEauthorblockA{Artificial Intelligence Institute\\
University of South Carolina\\
Columbia, SC, USA \\
\texttt{amit@sc.edu}}
}

\maketitle

\begin{abstract}
A fundamental question in natural language processing is - what kind of language structure and semantics is the language model capturing? Graph formats such as knowledge graphs are easy to evaluate as they explicitly express language semantics and structure. This study evaluates the semantics encoded in the self-attention transformers by leveraging explicit knowledge graph structures. We propose novel metrics to measure the reconstruction error when providing graph path sequences from a knowledge graph and trying to reproduce/reconstruct the same from the outputs of the self-attention transformer models. The opacity of language models has an immense bearing on societal issues of trust and explainable decision outcomes. Our findings suggest that language models are models of stochastic control processes for plausible language pattern generation. However, they do not ascribe object and concept-level meaning and semantics to the learned stochastic patterns such as those described in knowledge graphs. Furthermore, to enable robust evaluation of concept understanding by language models, we construct and make public an augmented language understanding benchmark built on the General Language Understanding Evaluation (GLUE) benchmark. This has significant application-level user trust implications as stochastic patterns without a strong sense of meaning cannot be trusted in high-stakes applications.
\end{abstract}


\section{Introduction}
Recent studies have studied self-attention models such as transformers for their ability to encode underlying graph structures by drawing parallels with graph neural networks (GNNs)\cite{choi2020learning} Intuitively, there is a correspondence between the self-attention map in the transformer and the normalized adjacency matrix in GNNs. Also, there is a correspondence between GNN node representations and the output value vectors from a transformer. The multiple routings of the transformer output through layers of the transformer are similar to multiple graph convolution aggregations in a GNN. Thus, both transformers may be an effective way to learn graph contexts between language tokens. In this study, we aim to test this perceived equivalence rigorously.

Do transformers encode semantic graphs between input sequence tokens? We perform simple experiments that feed various graph path sequence inputs to transformers (we test with multiple knowledge graphs (KGs) and language models (LMs)) and try reconstructing the input graph from transformer outputs. In our experiments, we find that in doing so, a high reconstruction error is observed for certain types of graph paths, \textit{paths that require strongly typed real-world concept level knowledge} (e.g., Volvo is typically a type of high-performance car which is, in turn, a type of car). Several previous works have performed similar knowledge graph-based reconstruction experiments. However, they have measured link prediction performance alone and not path predictions \cite{ma2023survey}. Link prediction is a weak evaluation of knowledge graph semantics as the richness of concepts in a knowledge graph comes from graph paths connecting concepts comprising multiple relationships. Furthermore, they have not qualitatively analyzed the results of successful and failed outcomes. In this study, we quantitatively measure the ability of transformers to predict relationships and concepts in knowledge graph paths. We also qualitatively inspect the paths on which the model makes errors to evaluate their conceptual understanding capabilities.
\section{Methdology}\label{sec:methods}
First, we extract masked graph paths from the knowledge graphs for processing by the language model. Figure \ref{fig:kpaths} illustrates the masked graph path extraction process from the knowledge graph. 
\begin{figure}[!htb]
    \centering
    \includegraphics[width=\linewidth,trim = 0cm 7cm 13cm 0cm, clip]{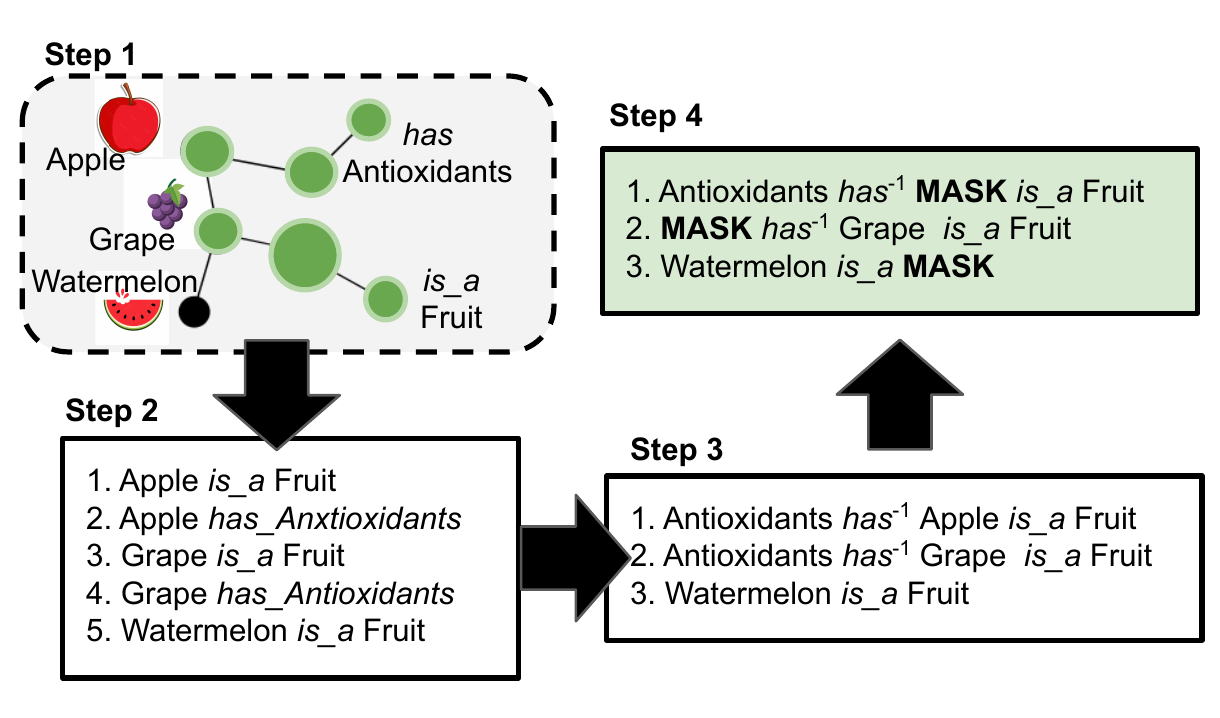}
    \caption{Steps 1, 2, and 3 show the process of converting the knowledge graph links to paths. Step 4 shows the masked inputs to the language model that will predict the masked tokens. The links are connected two make longer paths through the use of inverse relationships, e.g., \textit{has}\textsuperscript-\textsuperscript 1.}
    \label{fig:kpaths}
\end{figure}
Next, we predict the masked tokens and calculate the percentage of times the language models assign the correct token top five prediction ranks (measured using softmax over logits). Figure \ref{fig:path_pred} illustrates this process. The final softmax logits obtained can be ranked in order of probability values. For our evaluation metric, we calculate the percentage of times the correct answer is within the top five probabilities. We call this metric \textbf{\%Top@5}.
\begin{figure}[!htb]
    \centering
    \includegraphics[width=\linewidth,trim = 8cm 0cm 0cm 0cm, clip]{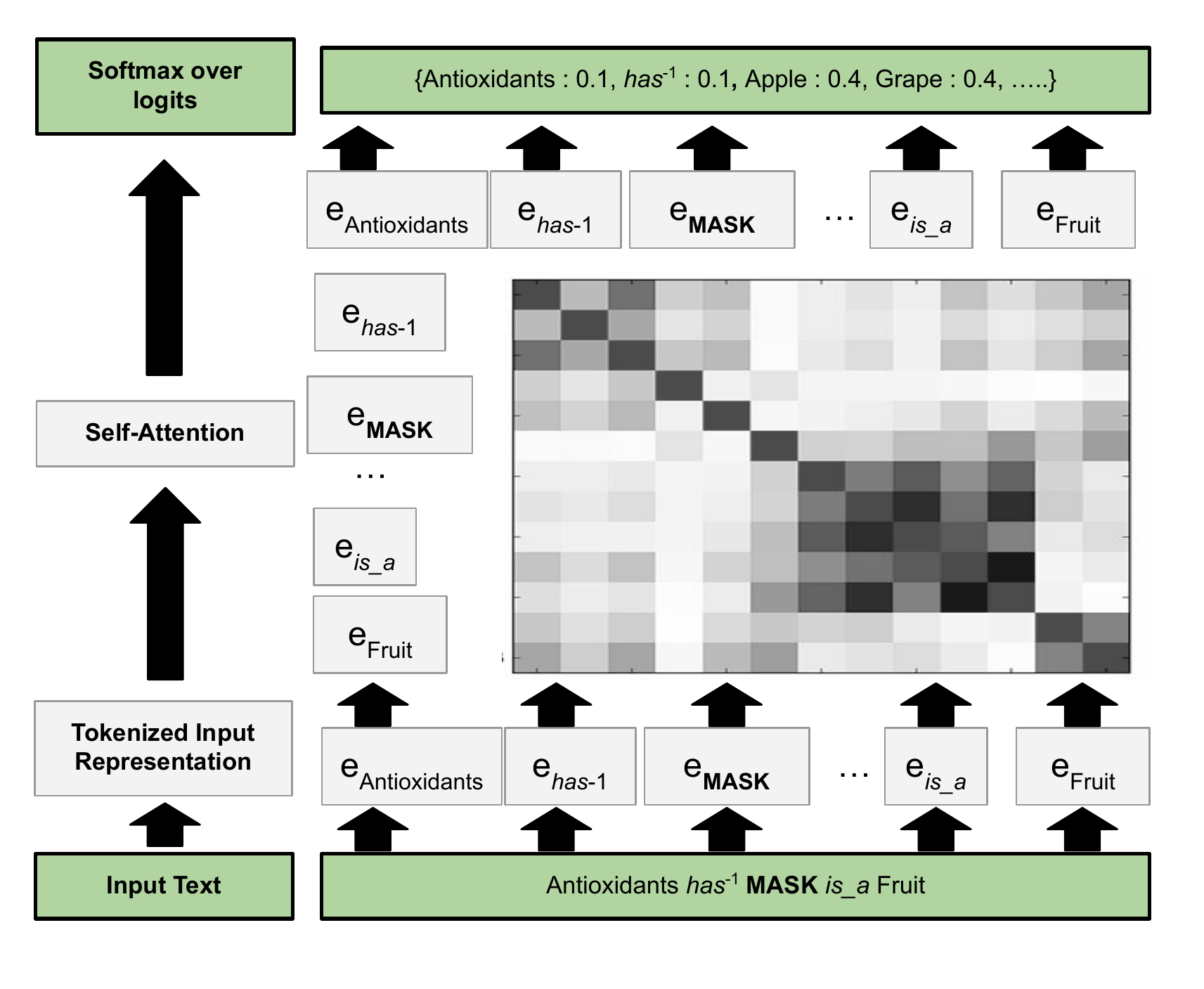}
    \caption{The figure shows how the masked graph path inputs are processed through the self-attention transformer models to obtain softmax logit outputs.}
    \label{fig:path_pred}
\end{figure}
\section{Experiments}\label{sec:exps}
We extract approximately 300K knowledge graph links from the knowledge graphs DBPedia, ConceptNet, Wiktionary, WordNet, and OpenCyc Ontology \cite{speer2017conceptnet}. The relationships we find are \textit{Antonym, DistinctFrom, EtymologicallyRelatedTo, LocatedNear, RelatedTo, SimilarTo, Synonym, AtLocation, CapableOf, Causes, CausesDesire, CreatedBy, DefinedAs, DerivedFrom, Desires, Entails, ExternalURL, FormOf, HasA, HasContext, HasFirstSubevent, HasLastSubevent, HasPrerequisite, HasProperty, InstanceOf, IsA, MadeOf, MannerOf, MotivatedByGoal, ObstructedBy, PartOf, ReceivesAction, SenseOf, SymbolOf, and UsedFor}. The data can be found at this \href{https://github.com/kauroy1994/Data_for_spring_2023/blob/master/data.zip}{link}. For the language models, we use bert-base-uncased, bert-large, GPT-Neo small, medium, and large with 0.1B, 0.3B, 1B, 2.7B, and 6B parameters, respectively (B stands for billion) \cite{gozalo2023chatgpt}.
\subsection{Quantitative Results}
Figure \ref{fig:exps} shows the quantitative results. We explain the results in the figure caption due to space limitations.
\begin{figure}[!htb]
    \centering
    \includegraphics[width=\linewidth,trim = 7cm 0cm 0cm 0cm, clip]{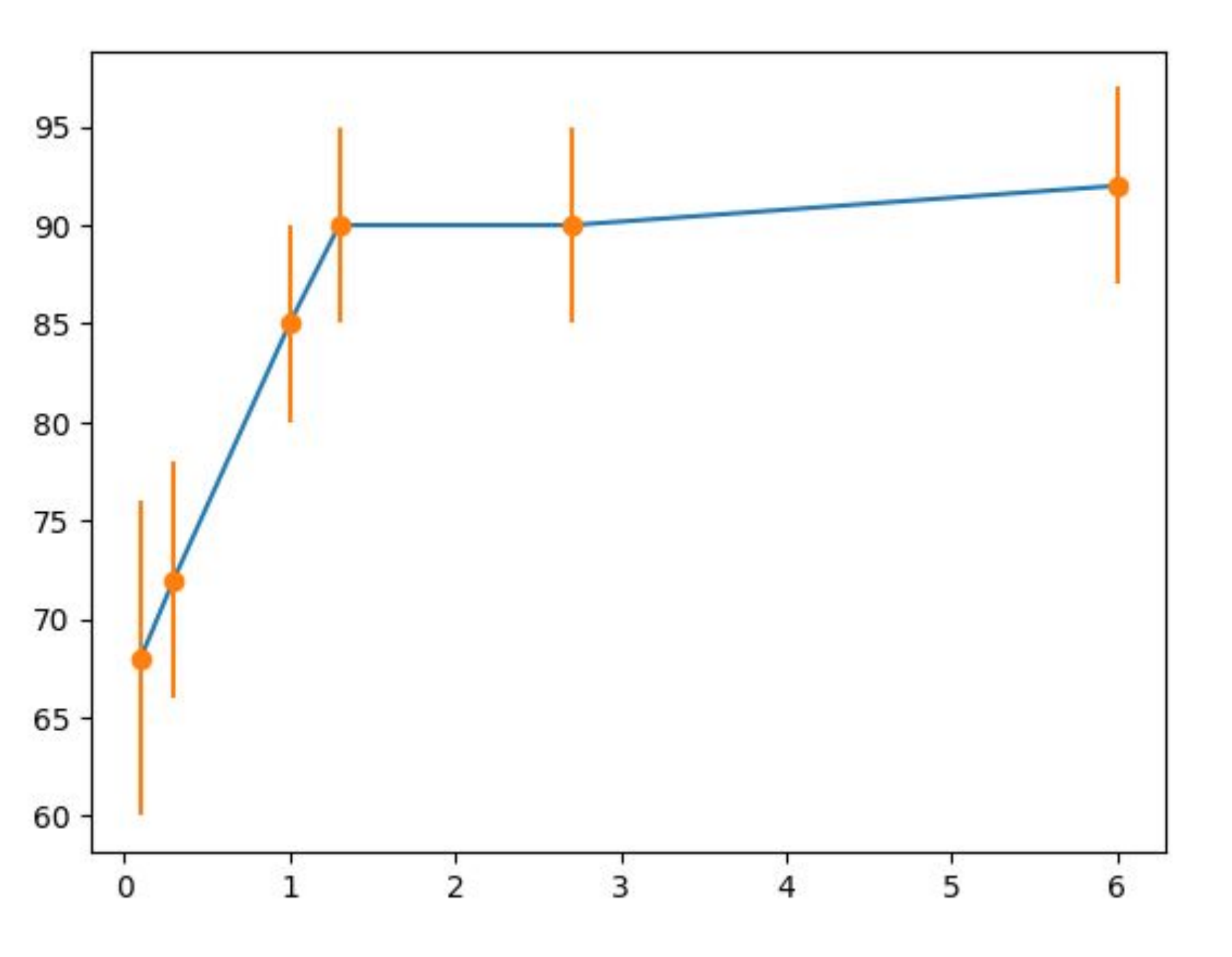}
    \caption{The X-axis denotes the number of parameters in billions, and the Y-axis measures the \textbf{\%Top@5}. The performance measured using \textbf{\%Top@5} increases steadily with the number of model parameters. However, after a certain amount of parameters is reached ($\sim$ 1 billion), the performance starts to flat-line. The variance across different runs remains significant ($\sim \pm$ 5), although it also shows a decreasing trend with increased model parameters. }
    \label{fig:exps}
\end{figure}
\subsection{Qualitative Results}
We manually inspect the knowledge graph paths at which the language models fail, which we will call \textit{false paths}. Interestingly, the \textit{false paths} almost exclusively involve knowledge of strongly typed objects and their properties as seen in the real world. Some examples include ``volvo IsA car CapableOf slow\_down'', ``retrograde\_motion HasContext astronomy IsA physics'', ``handicapped SimilarTo unfit RelatedTo unhealthy'', and ``ultimate\_frisbee IsA field\_game IsA outdoor\_game''. The remaining examples are at this \href{https://github.com/kauroy1994/Data_for_spring_2023/blob/master/false_paths.txt}{link}. This finding is particularly interesting as it supports third-party observations about language models' fundamental lack of a conceptual world model when asked about physics-related questions (e.g., block-stacking)\cite{bender2021dangers}. 

\section{The GLUE Benchmark}\label{sec:GLUE}
The General Language Understanding Evaluation (GLUE) benchmark revolutionized the evaluation of LMs with several challenging language understanding phenomena, specifically, Textual Entailment (TE), Textual Similarity (TS), and Natural Language Inference (NLI) \cite{wang2018glue}. The success of the GLUE benchmark directly led to rapid advances in language understanding evaluation benchmarks such as SuperGLUE, KILT, and BIG-BENCH, to name a few \cite{wang2019superglue,petroni2020kilt,srivastava2022beyond}. We identify defining characteristics of these benchmarks and explain why such characteristics are meaningless without further qualification by aligning with conceptual semantics such as those found in knowledge graphs. We first describe TE, TS, and NLI phenomena as they are currently established and defined. Then, we report shortcomings resulting from their incomplete definitions due to non-alignment with knowledge graph semantics.
\paragraph{Natural Language Inference (NLI)}The GLUE tasks - MNLI (Multi-genre Natural Language Inference), QNLI (Question Answering Natural Language Inference), and WNLI (Winograd Natural Language Inference) test NLI capabilities from varying angles. MNLI tests whether the model can appropriately judge if a sentence logically follows from another, i.e., logical entailment. QNLI tests similar logical entailment between question and statement pairs - does it make logical sense to ask a follow-up question? WNLI tests logical entailment in the presence of pronouns and the nouns they reference.
\paragraph{Textual Entailment (TE)} The GLUE task RTE (Recognizing Textual Entailment) tests for logical entailment similar to the NLI task MNLI. However, RTE emphasizes the meaning - given two text fragments, whether the meaning of one can be entailed (or can be inferred) from the other.
\paragraph{Textual Similarity (TS)}The GLUE task QQP (Quora Question Pairs) tests for the ability to assess the semantic equivalence, measured as the similarity between a pair of questions that appear on the social media forum Quora.  

Together, the TE, TS, and NLI phenomena constitute the fundamental \textit{``characteristics''} of language understanding. Many other characteristics define human-like language understanding, such as abstraction, analogy, and implicit mentions. We posit that if the TE, TS, and NLI characteristics are satisfactorily incorporated into an LM's basic skill set, those other characteristics can arise through compositions of this basic skill set.

\section{Constructing the Modified GLUE Benchmark}\label{sec:modglue}
\subsection{Knowledge Graphs Augmented GLUE} We now augment the GLUE benchmark using the Knowledge Graphs (KGs) DBPedia, ConceptNet, Wiktionary, WordNet, and the OpenCyc Ontology which consist of semantic associations as networks of connected objects and their relationships \cite{auer2007dbpedia,speer2017conceptnet,matuszek2006introduction}. These KGs are human-curated and carry ontological commitments (a good majority of humans that curated the knowledge agree on a common semantic interpretation of the KGs). For the first step of building the new benchmark, we augment the task datasets described in Section \ref{sec:GLUE}, using order one and order two (one-hop and two-hop paths) associations from the aforementioned knowledge sources. Tokenization of the data points is essential when querying knowledge sources for paths. Therefore, we define tokenization tailored to specific sources from which we extract knowledge paths based on manual inspection and determination of optimal practices and heuristics. For example, one of the tokenization techniques we use is using a sliding window on the input data and extracting tokens of span length 1-5 as we found entity mentions of span length 5 (e.g., Standard and Poor Index 500). We limit the augmentation to order two paths as we find higher-order paths introduce noisy concepts and paths to the dataset and can, therefore, hinder the semantic correctness of the dataset. 

The relationships we find are the same as described in Section \ref{sec:exps}. We make our augmented dataset available at this \href{https://drive.google.com/file/d/1nfY2DEZOc8dHpfwh5PLWkuTqCltq9U3l/view?usp=sharing}{link}.
\paragraph*{Concept Understanding Evaluation Metrics} Our augmented dataset supports the evaluation metric for knowledge graph path alignment, i.e., \textbf{\%Top@5}, introduced in Section \ref{sec:methods} as well as standard performance scores of accuracy, F1 score, precision, and recall, both being necessary to conclusively determine a language model's conceptual understanding capabilities.
\section{Conclusion}
This paper opens the black-box language models' ability to model knowledge graph semantics by proposing masked prediction tasks on graph paths. We do this to understand a language model's conceptual understanding and its bearings on application-level user trust issues. We introduce metrics for the evaluation of the results and also manually inspect the outcomes. 

Our findings suggest that language models are models of stochastic control processes for plausible language pattern generation. However, they do not ascribe object and concept-level meaning and semantics to the learned stochastic patterns such as those described in knowledge graphs. This has significant application-level user trust implications for applications requiring concept-level understanding (e.g., healthcare) and physical simulations (e.g., war-time strategies). Our findings suggest that using language models alone, which are stochastic control models, to drive high-stake application-level decisions would be highly unsafe and irresponsible. Finally, the paper constructs and makes public a knowledge-augmented GLUE benchmark that can foster the development of trustworthy language models through concept grounding in human-curated knowledge sources such as knowledge graphs. We also touch upon how novel evaluation metrics combined with such augmented datasets can be used to further increase the reliability and robustness of language model evaluations with regard to their ability to accurately encode conceptual understanding.
\section*{Acknowledgements}
This work was supported in part by the National Science Foundation under Grant 2133842, “EAGER: Advancing Neuro-symbolic AI with Deep Knowledge-infused Learning" and was carried out under the advisement of Prof. Amit Sheth \cite{sheth2023neurosymbolic,sheth2021knowledge}. 
\bibliographystyle{ieeetr}
\bibliography{refs}
\end{document}